\documentclass{article}

\usepackage{PRIMEarxiv}

\usepackage[utf8]{inputenc} % allow utf-8 input
\usepackage[T1]{fontenc}    % use 8-bit T1 fonts
\usepackage{hyperref}       % hyperlinks
\usepackage{url}            % simple URL typesetting
\usepackage{booktabs}       % professional-quality tables
\usepackage{amsfonts}       % blackboard math symbols
\usepackage{nicefrac}       % compact symbols for 1/2, etc.
\usepackage{microtype}      % microtypography
\usepackage{lipsum}
\usepackage{fancyhdr}       % header
\usepackage{graphicx}       % graphics
\graphicspath{{media/}}     % organize your images and other figures under media/ folder

\usepackage{algorithm} 
\usepackage{algorithmic}
\usepackage{adjustbox}
\usepackage{mathtools}

\newcommand{\CE}{\mathrm{CE}}

\usepackage[dvipsnames]{xcolor}

\title{\textcolor{Plum}{G-Drift} MIA: Membership Inference via \textcolor{Plum}{G}radient-Induced Feature \textcolor{Plum}{Drift} in LLMs
}

\author{
Ravi Ranjan\thanks{Corresponding author. \\Accepted to the \textbf{ICPR 2026} conference; to appear in the Springer LNCS proceedings. \\This version includes extended supplementary materials. Code: \url{https://github.com/raviranjan-ai/GDrift-ICPR-2026}}\\
Florida International University\\
Miami, USA\\
{\tt rkuma031@fiu.edu}
\and
Utkarsh Grover\\
University of South Florida\\
Tampa, USA \\
{\tt\small utkarshgrover@usf.edu}
\vspace{0.4cm}
\and
Xiaomin Lin\\
University of South Florida\\
Tampa, USA \\
{\tt\small xlin2@usf.edu}
\and
Agoritsa Polyzou\\
Florida International University\\
Miami, USA\\
{\tt\small apolyzou@fiu.edu}
}

\begin{document}
\maketitle

\begin{abstract}
Large language models (LLMs) are trained on massive web-scale corpora, raising growing concerns about privacy and copyright. Membership inference attacks (MIAs) aim to determine whether a given example was used during training. Existing LLM MIAs largely rely on output probabilities or loss values and often perform only marginally better than random guessing when members and non-members are drawn from the same distribution. 
We introduce \emph{G-Drift MIA}, a white-box membership inference method based on \emph{gradient-induced feature drift}. Given a candidate $(x,y)$, we apply a single targeted gradient-ascent step that increases its loss and measure the resulting changes in internal representations, including logits, hidden-layer activations, and projections onto fixed feature directions, before and after the update. These drift signals are used to train a lightweight logistic classifier that effectively separates members from non-members. Across multiple transformer-based LLMs and datasets derived from realistic MIA benchmarks, G-Drift substantially outperforms confidence-based, perplexity-based, and reference-based attacks. We further show that memorized training samples systematically exhibit smaller and more structured feature drift than non-members, providing a mechanistic link between gradient geometry, representation stability, and memorization. In general, our results demonstrate that small, controlled gradient interventions offer a practical tool for auditing the membership of training-data and assessing privacy risks in LLMs.
\end{abstract}
% keywords can be removed
% \keywords{First keyword \and Second keyword \and More}

\section{Introduction}
\label{sec:intro}
Large language models (LLMs) are trained on vast web-scale corpora, frequently collected without explicit consent from content creators. This practice has triggered growing legal, ethical, and regulatory scrutiny \cite{karamolegkou2023,mueller2024}, exemplified by recent high-profile copyright lawsuits against major LLM providers, such as \emph{New York Times v.\ OpenAI and Microsoft} \cite{freeman2024}.
A central question emerges from these debates: \emph{given a specific data instance, can we determine whether it was used to train an LLM?} This question formalizes the problem of distinguishing training samples (\textbf{members}) from unseen data (\textbf{non-members}), known as \emph{membership inference} \cite{Shokri2017}. In the context of large generative models, answering this question is critical for privacy auditing, transparency, and accountability, enabling data owners and regulators to verify claims about training data usage \cite{yan2024protecting}.

Membership inference attacks (MIAs) have a long history in machine learning \cite{Shokri2017,Yeom2018}, yet extending them to LLMs has proven surprisingly difficult \cite{aubinais2025membership,duan2024membership}. Most existing black-box MIAs rely on output-level signals, such as prediction confidence, likelihood, or loss, under the intuition that models are more confident on memorized examples. While this intuition holds weakly, recent studies demonstrate that such attacks are highly sensitive to distributional artifacts between member and non-member datasets \cite{Maini2024}. When evaluated under same-distribution conditions, many LLM MIAs degrade to near-random performance, highlighting the fundamental challenge of disentangling memorization from generalization in large generative models. 
Large-scale evaluations~\cite{hayesexploring} have shown that even strong reference-based attacks, such as LiRA \cite{carlini2022membership}, achieve suboptimal performance on GPT-2–style models under realistic Chinchilla-optimal training. These results suggest a fundamental ambiguity: either modern LLMs expose only weak and unstable membership signals, or existing attacks fail to probe the internal mechanisms where memorization is encoded. 

We resolve this ambiguity by introducing \textbf{G-Drift MIA}, a white-box membership inference attack based on \emph{gradient-induced feature drift}. We apply a single, controlled gradient-ascent step to locally increase the loss of a candidate example and measure the resulting changes in internal representations, including logits, hidden activations, and feature projections, to reveal robust membership signals. Our approach leverages the over-parameterized nature of LLMs and recent insights from mechanistic interpretability, which suggest that memorized samples occupy locally stable regions of representation space. Empirically, we observe that memorized training samples display consistently stronger and more structured feature drift than non-members, allowing reliable separation with a lightweight logistic classifier. Extensive experiments across transformer-based LLMs and realistic benchmarks demonstrate that G-Drift substantially outperforms confidence-based, perplexity-based, and reference-based MIAs.

In summary, our contributions are threefold: (i) we introduce gradient-induced feature drift as a principled and effective signal for white-box membership inference in LLMs; (ii) we demonstrate consistent and substantial performance gains over prior MIAs across models and datasets; and (iii) we provide empirical evidence linking representation stability under gradient perturbations to memorization. Together, these results establish controlled gradient interventions as a practical tool for auditing data-membership and assessing privacy risks in LLMs.

\section{Related Work}
\label{sec:related-work}

\subsection{Membership Inference Attacks (MIAs)}
Originally introduced by Shokri et al.~\cite{Shokri2017}, MIAs form the foundational framework for determining whether a data point was used during model training. Early research predominantly targeted classification models~\cite{huhongsheng2022,niu2024}, relying on the assumption that models assign higher confidence to their training samples. Classical black-box membership inference attacks rely on output probabilities or entropy-based confidence thresholds~\cite{Yeom2018}. Shadow modeling extends these ideas by training auxiliary models to approximate the target model’s behaviour under varying membership conditions~\cite{tonni2020data}, while ensemble-based and label-only formulations refine thresholding strategies under restricted access settings~\cite{Choquette2021}. Complementing these approaches, \textit{label-only} membership inference attacks operate without access to logits and instead depend solely on final predictions. PETAL~\cite{he2025towards} exemplifies this setting by leveraging token-level semantic similarity as a proxy for perplexity, demonstrating competitive performance among label-only strategies.

As MIAs transitioned to generative models such as LLMs, output-based methods revealed significant limitations. Carlini et al.~\cite{Carlini2022} demonstrated that membership detection for short GPT-2 sequences is ineffective unless the model strongly overfits. Further, Maini et al.~\cite{Maini2024} established that many previously reported gains were artifacts of temporal distribution shifts between member and non-member corpora; when these biases were eliminated, attack accuracy often collapsed to near-random performance~\cite{Elhage2022,meeus2025}. These findings highlight the fragility of output-only MIAs and underscore the need for richer internal signals.

Motivated by these challenges, recent research increasingly leverages internal model information through white-box MIAs. Perturbation-based approaches~\cite{shi2023membership}, neighborhood-comparison techniques~\cite{mattern2023membership}, and self-prompt calibration~\cite{fu2022spvmia} incorporate gradients or hidden-layer characteristics to improve membership detection. In CNNs and early deep networks, white-box attacks exploiting activations, gradients, or gradient-based memorization signals~\cite{Nasr2019,Leino2020,freeman2024} consistently outperformed black-box alternatives, revealing that internal-state access provides substantially stronger membership cues. Nasr et al.~\cite{Nasr2019} specifically showed that combining gradients with intermediate activations yields significant accuracy gains, while Leino and Fredrikson’s ``Stolen Memories''~\cite{Leino2020} operationalized memorization as the gradient effort required to ``forget'' a sample. Despite leveraging internal model states, existing white-box membership inference attacks remain imperfect, often achieving only modest accuracy
% with ROC–AUC values
\cite{hayesexploring}.

Complementary work studies memorization and extraction phenomena in LLMs, where models reproduce verbatim or near-verbatim segments of their training data~\cite{Carlini2022,tirumala2022memorization}, with the goal of characterizing and quantifying direct data leakage risks from generative outputs.
Membership inference differs crucially: the goal is not to reconstruct content but to detect whether a sample influenced training dynamics by contrasting behaviour across members and non-members~\cite{Shokri2017,carlini2022membership}. This distinction is important for understanding why MIAs require more subtle indicators than direct memorization leakage.

In the context of LLMs, a range of recent heuristics attempt to exploit model behaviour beyond raw likelihoods. Methods such as Min-$K\%$ probability~\cite{shi2023min_k}, Neighbor-MIA~\cite{mattern2023neighbormia}, and SPV-MIA~\cite{fu2022spvmia} rely on distributional cues or prompt-level perturbations, yet their effectiveness is often inconsistent across domains and architectures. A more structured methodology is offered by LUMIA~\cite{ibanez2025lumia}, which employs linear probes over intermediate transformer representations to infer membership. This line of work underscores a key observation: membership signals are distributed unevenly across layers, with deeper representations frequently exhibiting stronger separability.

Our method contributes to this growing direction of \emph{looking inside the model} and addresses the shortcomings of both black-box and probe-based white-box MIAs. Rather than analyzing static internal representations, we apply a targeted gradient-ascent perturbation that slightly ``unlearns'' a candidate sample and measure the induced \emph{gradient-induced feature drift}.
Unlike LUMIA, which trains many probes across layers, we rely on a single, lightweight gradient intervention and evaluate the model’s immediate reaction: \emph{If we nudge the model away from this sample, how strongly does it resist?} This produces a compact, robust, and theoretically grounded white-box signal for membership inference.

\subsection{Mechanistic Interpretability}
Our use of internal feature directions builds on advances in mechanistic interpretability that examine how high-dimensional representations encode knowledge in LLMs. Elhage et~al.~\cite{Elhage2022} demonstrated that neural networks frequently represent multiple concepts within a single neuron through \emph{polysemanticity}, meaning that activations often correspond to superpositions of unrelated features. This phenomenon complicates attempts to directly interpret individual activations or neurons. To address this challenge, Sparse Autoencoders (SAEs) have been proposed as a means to disentangle overlapping features and uncover more monosemantic, interpretable directions in activation space~\cite{Bricken2023}.

Moving beyond static analysis, Xu et~al.~\cite{xu2024tracking} introduced \emph{SAE-Track}, a method for tracing how interpretable features emerge, stabilize, and evolve across model training checkpoints. Their findings offer insight into how knowledge, including memorized information, is gradually embedded into internal representations. 
Gong et~al.~\cite{Gong2025} explored the security implications of polysemantic features, enabling adversarial manipulation of model behaviour through targeted interventions in representation space.
G-Drift leverages this interpretability perspective for \emph{diagnosis} rather than manipulation. 
We exploit representation geometry to measure how internal features \emph{shift} under a controlled gradient perturbation. This shift, captured as gradient-induced feature drift, provides a principled signal for identifying whether a sample was memorized during training.

\section{Methodology}
\label{sec-method}
\subsection{Problem Setup}
We assume a target LLM with parameters $\theta$ (e.g., the weights of a multi-layer transformer), which has been trained on some dataset $D_{\text{train}}$. We do not initially know whether a particular sample $x$ (with its ground-truth output $y$) was in $D_{\text{train}}$. Our goal is to construct a \textbf{membership classifier} $\mathcal{M}$ that outputs a high probability if $(x,y) \in D_{\text{train}}$ (member) and low if not (non-member). We operate in a \emph{white-box} scenario: we have access to $\theta$ and can perform forward and backward passes through the model. This setting is plausible in scenarios like a company auditing its own model or a regulator inspecting a model with cooperation from its owner (but without knowing the training data a priori). 

\subsection{Intuition}
Our approach is motivated by the observation that a small \emph{gradient-ascent} step used as a controlled ``unlearning'' perturbation elicits systematically different internal responses for memorized (member) versus unseen (non-member) samples, reflecting differences in how these examples are encoded in the model’s loss landscape and representation geometry. For a member sample $(x,y)\in D_{\text{train}}$, training actively shapes the model parameters around $(x,y)$, creating localized, content-specific representations. Consequently, even a single ascent update produces a \emph{measurable} 
change in the internal representation of member instances that is more noticeable compared to non-member instances.

At the same time, we expect that any gradient ascent step will alter the representations of non-members in a more unstructured and random way than for non-members that are harder to unlearn in a single step. 
The next section formalizes this intuition by defining ``feature projections'' and quantifying their induced drift for membership classification.

\subsection{Proposed approach: G-Drift MIA}

\begin{figure}[t]
  \centering
  % adjust width as needed; here we use \textwidth
  \includegraphics[width=0.9\textwidth]{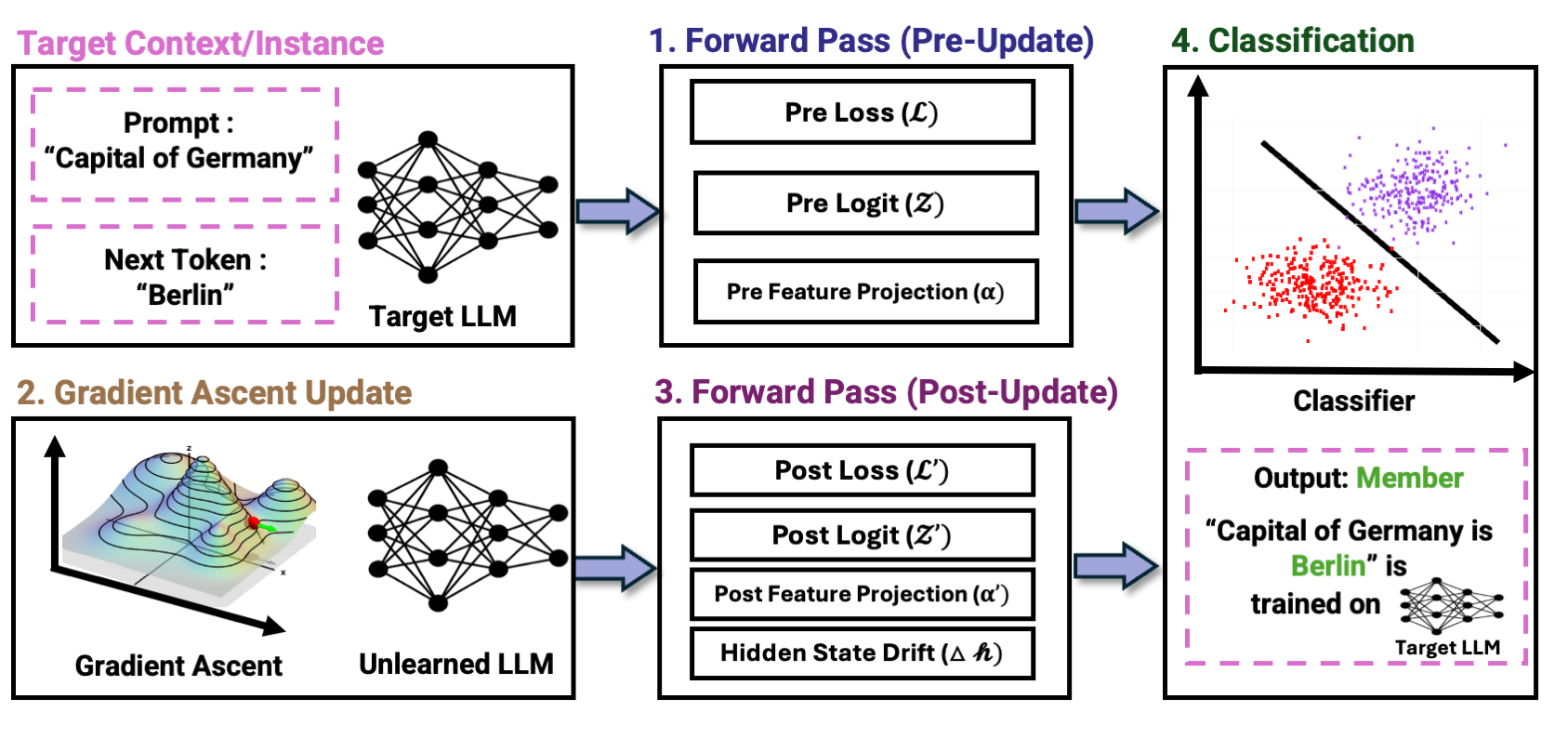}
  \caption{Approach overview: Gradient-Induced Feature Drift (G-Drift) Attack.}
  \label{fig:our-approach}
\end{figure}

Our approach consists of measuring how the model's predictions and internal representations for $(x,y)$ change when performing a single \emph{adversarial parameter update}, aiming to increase the loss of $(x,y)$. We call this change \textbf{feature drift}. Figure \ref{fig:our-approach} shows our proposed approach, G-Drift MIA, where we perform a forward pass of $x$ to our target LLM, a gradient ascent nudge, and a post-nudge pass of $x$, while collecting relevant features. We then use the vector of these features, $\mathbf{f}_x$, to perform the final membership classification of $x$ using a logistic regression model.
For a given sample $x$, the steps of G-Drift MIA are as follows:

\begin{enumerate}
    % Forward Pass (Pre-Update)
    \item \textbf{Forward Pass (Pre-Update):} 
    We input the prompt \(x\) into the target LLM model, parameterized by \(\theta\), $f_\theta(x)$, typically an auto-regressive transformer, to predict the next token \(y\). The model outputs a logit vector \(z = f_\theta(x) \in \mathbb{R}^V\), where \(V\) is the vocabulary size, and $z$ represents the unnormalized scores for each possible token. We also record the final-layer hidden state \(h \in \mathbb{R}^d\), such as the residual stream activation at the position preceding \(y\). This hidden state encapsulates the model's internal representation of the context. We compute the following features. 

    % Cross-Entropy Loss
    \textbf{(a) Pre-Update Loss ($\mathcal{L}$):} 
    We compute the cross-entropy loss: 
    \[\mathcal{L}\bigl(\theta;x,y\bigr) = -\log \bigl[\mathrm{softmax}\bigl(f_{\theta}(x)\bigr)_y\bigr].\]
    This loss quantifies the model's confidence in predicting \(y\), with lower values indicating higher confidence. It serves as a basic feature
    % baseline metric
    for assessing the model's pre-update performance on \((x, y)\).

    % Pre-Update Logit
    \textbf{(b) Pre-Update Logit $(z_y)$ :} 
    We extract \(z_y\), the logit corresponding to the true token \(y\) from the vector \(z\). This raw score reflects the model's unnormalized confidence in \(y\) before the softmax normalization, providing a direct measure of prediction strength for comparison post-update.

    % Feature Projection (Pre-Update)
    \textbf{(c) Pre-Update Feature Projection (${\alpha}$):} 
    To probe specific aspects of the model's internal representation, we select a direction vector \(v \in \mathbb{R}^d\) in the hidden state space, matching the dimensionality of \(h\). In the simplest case, \(v\) is a random unit vector, though more informed choices (e.g., directions learned via probing for specific concepts)
    can enhance interpretability. We compute the scalar projection \(a = h^\top v\), which measures the alignment of \(h\) with \(v\), effectively quantifying the activation of the feature represented by \(v\).

    % Gradient Ascent Update
    \item \textbf{Gradient Ascent Update:} 
    We compute the gradient of the cross-entropy loss with respect to the model parameters, \(\nabla_\theta \mathcal{L}_{\CE}(\theta; x, y)\), and perform a single-step update in the direction of increasing loss:
    \[
    \theta' \leftarrow \theta + \eta \nabla_\theta \mathcal{L}_{\CE}(\theta; x, y),
    \]
    where \(\eta\) is the learning rate (e.g., \(10^{-2}\)). This adversarial update, contrary to standard gradient descent, intentionally degrades the model's performance on \((x, y)\), simulating an ``unlearning'' process. A small \(\eta\) ensures measurable changes without destabilizing the model, allowing us to study the impact of parameter perturbations.

    % Forward Pass (Post-Update)
    \item \textbf{Forward Pass (Post-Update):} 
    Using the updated parameters \(\theta'\), we reprocess the input \(x\) to obtain new logits \(z' = f_{\theta'}(x) \in \mathbb{R}^V\) and a new final-layer hidden state \(h' \in \mathbb{R}^d\). This step captures the model's altered behavior after the adversarial update, enabling a comparison with the pre-update state. We compute again the following changed feature values. 

    % Post-Update Logit
    \textbf{(a) Post-Update Loss ($\mathcal{L'}$) :} 
    After performing the single‐step ``unlearning'', we compute the cross‐entropy loss the same way as in Pre-Updated Loss. 

    % Post-Update Logit
    \textbf{(b) Post-Update Logit $(z'_y)$:} 
    We record \(z'_y\), the logit for the correct token \(y\) in \(z'\). When comparing \(z'_y\) with \(z_y\), we observe the change in the model's confidence in predicting \(y\), reflecting the effect of the gradient ascent update.

    % Feature Projection (Post-Update)
    \textbf{(c) Post-Update Feature Projection (${\alpha'}$):} 
    We compute \(\alpha' = {h'}^\top v\), the projection of the new hidden state \(h'\) onto the same direction \(v\). The difference between \(\alpha\) and \(\alpha'\) indicates how the adversarial update affects the activation of the feature represented by \(v\), providing insight into changes in the model's internal representation.

    % Hidden State Drift
    \textbf{(d) Hidden State Drift ($\Delta h$):} 
    We measure the Euclidean distance between the pre- and post-update hidden states, \(\Delta h = {||h' - h||}\). This metric, termed \emph{hidden state drift}, quantifies the overall shift in the model's internal representation due to the update, capturing the broader impact beyond specific feature directions.
    
    \item \textbf{Classification:} \label{Log-Reg_method}
    We use logistic regression to classify instance $(x,y)$ based on the three before- and four after-update features, including \(\Delta h\), collected in steps 1 and 3. Instance $x$ is represented as the following feature vector:
    \begin{equation}
    \textbf{f}_x =  \bigl[\mathcal{L},z_y ,\alpha, \mathcal{L'},z'_y ,\alpha' ,\Delta h \bigr].
    \end{equation}
    This feature vector characterizes the effect of the gradient intervention on the model for sample $x$. The classifier, $\mathcal{M}(\textbf{f}_x)$, outputs a probability reflecting membership likelihood. Using a threshold, we can generate binary predictions and balance precision and recall as required.  
\end{enumerate}

\subsection{Choice of Feature Direction}
\label{sub:feature-direction}
The direction \( v \) for projecting hidden states could be viewed as a hyperparameter in our method. While multiple directions or the full hidden state vector could be employed, a single random unit vector \( v \) suffices to detect significant differences between \( h \) and \( h' \), where \( \Delta a = (h' - h)^\top v \) captures the drift along a random axis~\cite{xu2024tracking,hinder2022suitability}.  Alternatively, \( v \) can be selected strategically: one option is to align \( v \) with \( h \) to assess norm variations, while another involves using advanced techniques like SAE-Track \cite{xu2024tracking} for sparse feature extraction to identify key directional bases, such as monosemantic vectors tied to specific concepts. Projecting onto such vectors might reveal membership signatures, particularly for content-relevant features in \(x\). Our approach leverages the likelihood that a random projection will reflect substantial hidden state changes in any direction. Employing additional directions could enhance the characterization of drift at the expense of increased complexity. So, our experiments utilize a fixed random $v$ for simplicity and consistency. 

The detailed pseudo-code for the proposed algorithm is provided in Appendix~\ref{app:pseudo}.

\section{Experimental Setup}
\label{sec:experiment-setup}

\subsection{Target LLM Models}
We evaluate G-Drift across three transformer-based LLMs to assess robustness across architectures and training regimes while maintaining comparable model scale.
\textbf{LLaMA-3.2 1B} is a compact variant of Meta’s LLaMA family \cite{Touvron2023}, containing approximately 1.3 billion parameters and serving as our primary evaluation model due to its strong memorization capacity despite modest scale.
\textbf{GPT-Neo 2.7B} is an open-source autoregressive transformer designed to replicate GPT-3-style training and architecture \cite{GPTNeo2021}. Its widespread adoption makes it a representative baseline for evaluating membership inference under realistic deployment conditions.
\textbf{Gemma-3 4B} is Google’s lightweight yet high-performance open LLM~\cite{team2024gemma}, trained with modern instruction tuning and alignment techniques. Its inclusion allows us to assess the generality of gradient-induced feature drift across independently developed architectures and training pipelines.

\subsection{Datasets (Members and Non-Members)}
Constructing reliable member–non-member datasets for LLM MIAs requires that both classes come from the same underlying distribution to avoid trivial artifacts, a concern emphasized in recent MIA and unlearning work. We therefore use datasets where each instance is explicitly labeled as \textit{member} or \textit{non-member} and where both share the same textual format and domain. Using only the member instances, we fine-tune the target LLMs to induce strong memorization of the corresponding facts. For all three datasets, non-members are drawn from the \emph{same} underlying distributions as the member instances, using two mechanisms: (i) \emph{future facts}, where questions are paired with correct answers about events or entities outside the model’s pre-training window, and (ii) \emph{counterfactual options}, where answers are replaced with plausible but incorrect entities, ensuring the specific (question, answer) pair was not seen during fine-tuning.

We conduct experiments on three commonly used datasets that have a Question-and-Answer (Q\&A) format. From the original datasets, we sample 1000 instances (half members and half non-members).
\textbf{WikiMIA}~\cite{shi2023detecting} provides Wikipedia-style factual content, which we convert into Q\&A pairs (e.g., ``Q: What is the capital of France? A: Paris''). \textbf{World Facts}~\cite{maini2024tofu} supplies  factual Q\&A instances with a similar structure.
\textbf{Real Authors 3}~\cite{maini2024tofu} consists of biography-style Q\&A about real authors, following the TOFU format. 

\subsection{Data Splits}
For each dataset, we construct balanced member and non-member sets and partition them into 70\% training, 10\% validation, and 20\% test splits. Each split maintains an equal number of member and non-member examples to avoid class-imbalance artifacts in the attack classifier. Hyperparameters are selected on the validation set, and final performance is reported on the held-out test set using True Positive Rate (TPR), False Positive Rate (FPR), and Area Under the Receiver Operating Characteristic (ROC) curve (ROC-AUC, or simply AUC).
 
\subsection{Competing Approaches}
We compare G-Drift against a suite of strong membership inference baselines spanning black-box, reference-based, and label-only settings.

\textbf{Shadow Model Attack (Neighbour-MIA).} Neighbour-MIA \cite{mattern2023membership} trains shadow models on auxiliary data to mimic the target, then infers membership by comparing the queried sample’s outputs to those seen during shadow training, representing the classical shadow-model–based threat model. \textbf{Similarity-based Attack (SPV-MIA).}
SPV-MIA \cite{fu2024selfprompt} assumes access to reference data from the same distribution and estimates membership by comparing the similarity between the target model’s output on the query and its outputs on the reference set, calibrating a decision threshold using known non-members. \textbf{PETAL (Label-Only MIA).} This model operates in a stricter label-only setting~\cite{he2025towards}. It leverages semantic similarity between model-predicted tokens and reference answers as a proxy for likelihood, and serves as a competitive baseline when logits are unavailable.
\textbf{Probability Threshold (Min-k\%).}
This black-box attack uses the model’s prediction confidence for membership inference \cite{shi2023min_k}, computing the percentile rank of the true likelihood $p(y\mid x)$ within a reference distribution and predicting membership when it exceeds a $k^\text{th}$-percentile threshold.
\textbf{Perplexity-Based Likelihood (Perplexity-PL) and Zlib.}
Following \cite{Carlini2022}, Perplexity-PL treats sequence-level perplexity (negative log-likelihood) as the test statistic, while the Zlib variant combines perplexity with a compression-based score to flag unusually easy-to-predict or compressible samples as members. Please refer to Appendix~\ref{app:exp-setting} for the additional experimental settings.

\section{Results}
\label{sec-results}

\begin{table}[b]
    \centering
    \small
    \caption{Main G-Drift membership inference results (AUC $\uparrow$) across datasets and LLMs. 
    Best value in each column is shown in bold.}
    \label{tab:gdrift_main_auc}
    \setlength{\tabcolsep}{4pt}
    \renewcommand{\arraystretch}{1.1}
    \begin{adjustbox}{max width=\textwidth}
    \begin{tabular}{lccccccccc}
        \toprule
        & \multicolumn{3}{c}{WikiMIA} 
        & \multicolumn{3}{c}{World Facts} 
        & \multicolumn{3}{c}{Real Authors} \\
        \cmidrule(lr){2-4} \cmidrule(lr){5-7} \cmidrule(lr){8-10}
        Methods 
        & Llama-3 & Gemma-3 & GPT-Neo-2 
        & Llama-3 & Gemma-3 & GPT-Neo-2 
        & Llama-3 & Gemma-3 & GPT-Neo-2 \\
        \midrule
        Neighbour-MIA \cite{mattern2023membership}
        & 0.6187 & 0.6092 & 0.6488 
        & 0.6040 & 0.5802 & 0.6518 
        & 0.6144 & 0.6490 & 0.6542 \\
        SPV-MIA \cite{fu2024selfprompt}
        & 0.4431 & 0.4642 & 0.4232 
        & 0.4022 & 0.4522 & 0.4142 
        & 0.4401 & 0.4844 & 0.4482 \\
        PETAL \cite{he2025towards}
        & 0.5200 & 0.5600 & 0.5600 
        & 0.5700 & 0.5880 & 0.5800 
        & 0.5600 & 0.5800 & 0.5800 \\
        Min-k\% \cite{shi2023min_k}
        & 0.8444 & 0.8454 & 0.8474 
        & 0.8244 & \textbf{0.8998} & 0.8244 
        & 0.8328 & 0.8654 & 0.8820 \\
        Perplexity-PL \cite{Carlini2022}
        & 0.5020 & 0.5580 & 0.6200 
        & 0.6122 & 0.5900 & 0.6220 
        & 0.5430 & 0.5820 & 0.6120 \\
        Zlib \cite{Carlini2022}
        & 0.5210 & 0.5320 & 0.5800 
        & 0.5120 & 0.4900 & 0.5610 
        & 0.5200 & 0.5410 & 0.5182 \\
        G-Drift (ours) 
        & \textbf{0.9600} & \textbf{0.9990} & \textbf{0.9998} 
        & \textbf{0.9392} & 0.8997 & \textbf{0.9983} 
        & \textbf{0.9375} & \textbf{0.9950} & \textbf{0.9350} \\
        \bottomrule
    \end{tabular}
    \end{adjustbox}
\end{table}

\subsection{Overall Attack Performance}
We first compare (in terms of AUC) our proposed method against six other competing approaches in Table~\ref{tab:gdrift_main_auc}. The results show that \textbf{G-Drift consistently outperforms all existing MIA baselines} across datasets and language models, while also aligning with performance trends reported in prior work. As expected, {ZLib}, {SPV-MIA}, and {PETAL} remain the weakest methods, with AUC values hovering close to $0.5$, indicating near-random behavior across all models. We also observe that {Perplexity-based (PPL) attacks exhibit higher variance} compared to Neighbour-MIA, whose scores remain relatively stable across settings. Among the competing approaches,Min-$k\%$ emerges as the second-best method overall, even achieving the top score in one configuration (with G-Drift closely behind), yet still falling notably short of the much higher and more consistent performance of G-Drift.

\begin{figure}[bt]
\centering
\includegraphics[width=0.95\columnwidth]{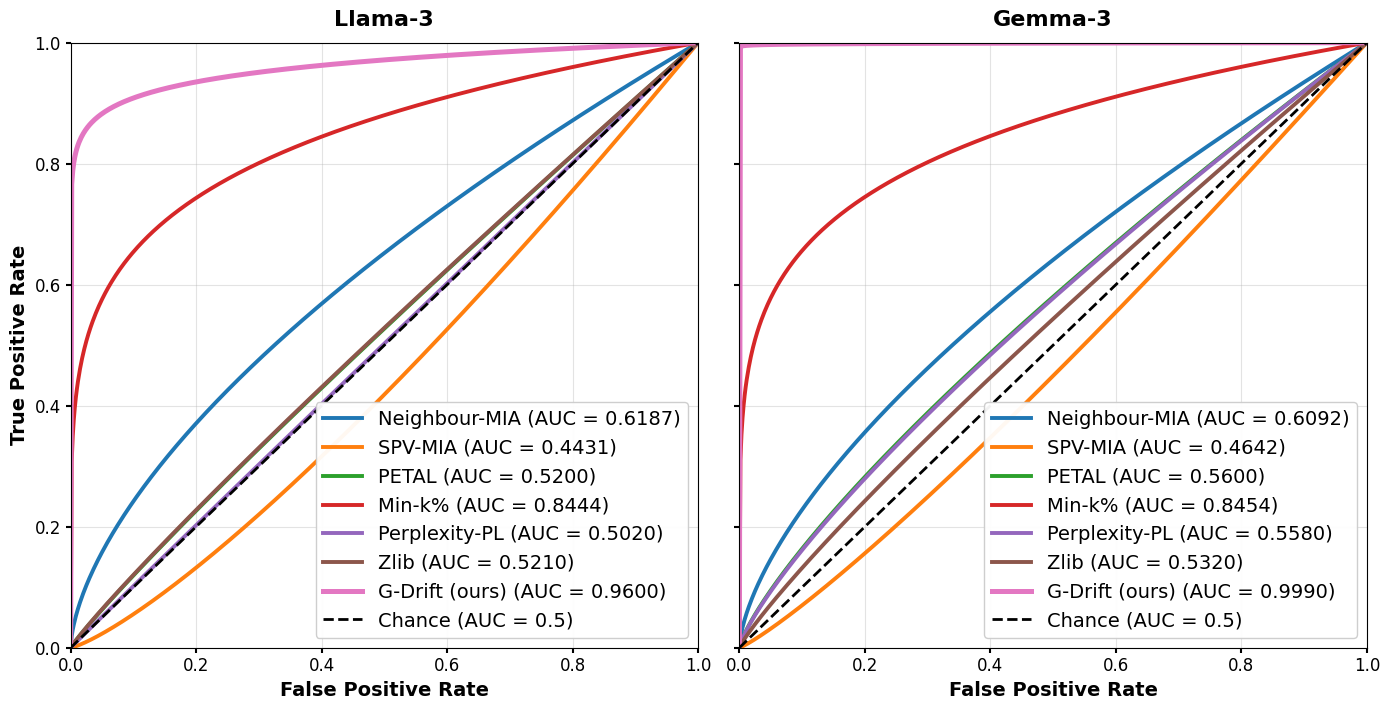} 
\caption{Comparison of ROC of Llama-3 \& Gemma-3 models on WikiMIA dataset}
\label{Roc-Curve-Dia}
\end{figure}

Figure~\ref{Roc-Curve-Dia} shows the full ROC for Llama-3 and Gemma-3 models evaluated on the WikiMIA dataset. G-Drift achieves an area under the curve of $96\%$ (on Llama-3) and $99.90\%$ (on Gemma-3), as its ROC curve is close to the ideal operating point at $(0,1)$, indicating strong separability between members and non-members. In contrast, Neighbour-MIA and SPV-MIA remain close to the diagonal, reflecting performance that is similar to random guessing.

\subsection{Ablation Study}
\label{sec:ablation}

\begin{table}[t]
\centering
\small
\setlength{\tabcolsep}{6pt}
\caption{Ablation study on WikiMIA dataset across three LLMs. 
Each row removes one or more features from G-Drift. Full feature set (“all”) provides the highest AUC. The lower the AUC, the more significant the feature(s) removed.}
\begin{tabular}{lccc}
\toprule
\textbf{Feature Set} & \textbf{Llama-3} & \textbf{GPT-Neo-2} & \textbf{Gemma-3} \\
\midrule
all & 0.9600 & 0.9998 & 0.9990 \\
all but loss before & 0.9597 & 0.9996 & 0.9987 \\
all but logit before & 0.9598 & 0.9997 & 0.9988 \\
all but feat proj before & 0.9596 & 0.9996 & 0.9986 \\
all but loss after & 0.9597 & 0.9996 & 0.9987 \\
all but logit after & 0.9598 & 0.9997 & 0.9988 \\
all but feat proj after & 0.9596 & 0.9996 & 0.9986 \\
\midrule
all but euclid drift & 0.9508 & 0.9912 & 0.9920 \\
\midrule
all but group before & 0.9342 & 0.9610 & 0.9686 \\
all but group after & 0.9502 & 0.9882 & 0.9910 \\
\midrule
all but loss & 0.9462 & 0.9800 & 0.9782 \\
all but logit & 0.9468 & 0.9812 & 0.9790 \\
all but feat proj & 0.9040 & 0.9342 & 0.9382 \\
\bottomrule
\end{tabular}
\label{tab:gdrift_ablation}
\end{table}

Table~\ref{tab:gdrift_ablation} reports an ablation analysis that quantifies the contribution of individual feature groups to G-Drift’s membership inference performance, measured by AUC on the WikiMIA dataset across three LLMs. We observe how our model performs when removing one or more features from $\mathbf{f}_x$. In general, removing informative features consistently degrades performance, with larger drops indicating a stronger reliance on the corresponding signals.

We first perform a leave-one-out analysis (rows 2 to 9), removing a single feature at a time. While doing that, we do not observe a sharp drop in AUC, suggesting that the feature groups carry partially overlapping membership information. 
Among single-feature variants, the strongest standalone signal (with the sharpest decline in performance once removed) is the Euclidean hidden-state drift, which captures a broader, more general summary of the model's behavior before and after the gradient-ascent nudge.

We then study the temporal structure of features (row 10 \& 11) by ablating an entire group of features collected before or after the gradient update. Removing the \emph{before-update} feature group leads to a substantial performance decline (e.g., Llama-3 drops from 0.9600 to 0.9342), whereas removing only the after-update group causes a noticeably smaller reduction. This indicates that most membership information is already present in the model’s original state, while post-update features primarily refine the decision boundary.

Finally, jointly removing the same feature type (row 12 to 14), both before and after the update, reveals that eliminating feature projections yields the most severe degradation across all models (down to 0.9040 on Llama-3), reinforcing their central role in G-Drift. Together, these results show that while all components contribute towards membership prediction, feature projections are the most critical signals driving membership inference performance.

\subsection{Analysis of Features' Drift}

\begin{figure}[bt]
  \centering
  \includegraphics[height=4.6cm, width=\textwidth]{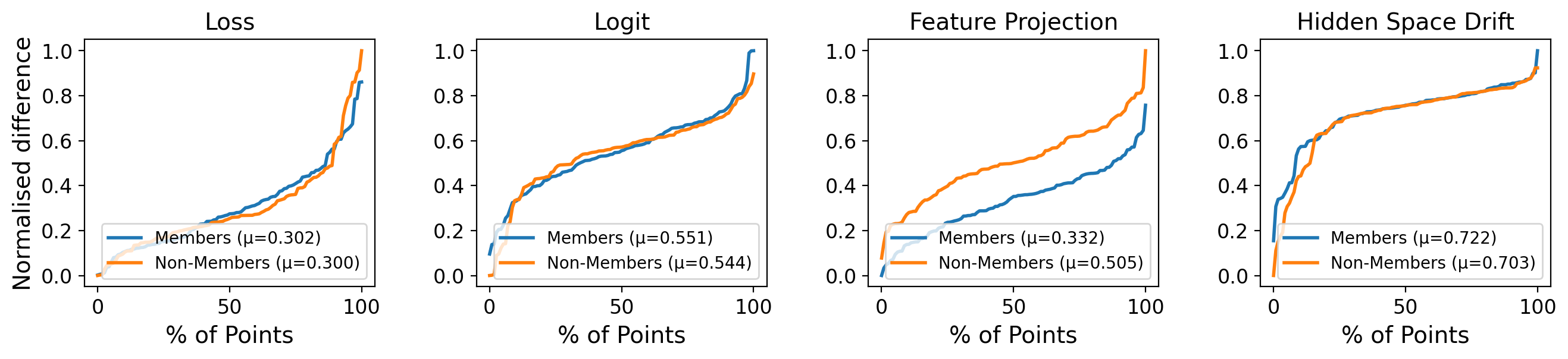}
  \caption{Analysis of min-max normalized features drift ($(\mathcal{L}'-\mathcal{L})$, $(z'_y-z_y), (\alpha'-\alpha), \Delta h$ respectively) with Llama-3 model on the WikiMIA dataset}
  \label{fig:Analysis}
\end{figure}

Figure~\ref{fig:Analysis} plots the cumulative distributions of the \emph{normalized} changes induced by a single gradient-ascent nudge for four signals: loss $(\mathcal{L}'-\mathcal{L})$, logit $(z'_y-z_y)$, feature projection $(\alpha'-\alpha)$, and hidden-space drift ($\Delta h$). 
On average, members exhibit marginally larger changes in loss and logits (members: $\mu{=}0.302$ vs.\ non-members: $\mu{=}0.300$ for loss; members: $\mu{=}0.551$ vs.\ non-members: $\mu{=}0.544$ for logits), and also a slightly higher hidden drift, which captures the magnitude of representational change (members: $\mu{=}0.722$ vs.\ non-members: $\mu{=}0.703$).

More interestingly, feature projection for members shows \emph{lower} differences on average (members: $\mu{=}0.332$ vs.\ non-members: $\mu{=}0.505$), yielding a clearer separation between the two classes. $\Delta \alpha$ measures alignment with a random direction, which we expect to be higher for non-members after the gradient ascent update, since they are not as well anchored as the member instances.
The figure suggests that, while several drift signals change comparably across classes, the feature-projection drift provides the most distinctive membership cue in our setting.
Overall, gradient-induced perturbations disproportionately affect internal representations associated with training (member) data, with feature projection drift providing the most discriminative signal.

A comprehensive quantitative analysis of the G-Drift framework, including consistency evaluations across semantically equivalent prompts, is detailed in Appendix~\ref{app:quant}.

\section{Discussion}
Our results suggest that membership signals in LLMs are more effectively exposed through \emph{controlled internal perturbations} than through output confidence alone: by applying a single gradient-ascent step and measuring the induced changes in logits, hidden states, and especially feature projections, G\textsc{-}Drift reveals a stronger mechanistic link between memorization, representation geometry, and feature-level stability. This perspective also aligns naturally with recent interpretability work on feature superposition and feature dynamics, which views model knowledge as encoded in evolving directions of representation space rather than in isolated neurons. At the same time, G\textsc{-}Drift remains a white-box method and its current evaluation is limited to specific datasets and Q\&A-style settings, so broader validation across architectures, modalities, and privacy-preserving training regimes remains an important direction for future work.

\subsection{Limitations}
Although G-Drift demonstrates strong membership inference performance, it has several important limitations. First, it is inherently a \emph{white-box} method that requires access to model parameters and gradients, and is therefore not applicable when only black-box query access is available~\cite{Carlini2022,Shokri2017}.
Second, despite our careful experimental design, our evaluation is limited to specific datasets and Q\&A-style formats. Broader empirical validation is needed to assess how well gradient-induced feature drift generalizes across diverse architectures, training regimes, data modalities, and preprocessing pipelines~\cite{fu2024selfprompt,hu2022membership}.
Finally, differential privacy (DP) provides a principled defense against memorization~\cite{abadi2016deep}. In DP-trained models, gradients for individual samples are intentionally noisy, reducing the separability between members and non-members in gradient-derived statistics and thereby weakening the effectiveness of our attack~\cite{carlini2022membership,Nasr2019}.

\subsection{Future Work}
Future work can extend G-Drift in several directions. One promising avenue is to replace random feature directions with more informative axes identified via interpretability methods, enabling stronger and more targeted drift signals. G-Drift can also be generalized to multi-modal models, where membership may be reflected in joint text–image representations. Finally, integrating G-Drift with dataset-level inference frameworks could allow aggregated drift evidence across many samples to support large-scale auditing tasks.

\section{Conclusion}
\label{sec-conclusion}
We introduce a novel membership inference attack on large language models using a single-step gradient ascent update. Our work combines the most significant insights from the existing works on MIA in LLMs, as well as other related works, i.e., superposition, interpretable feature directions, and unlearning in LLMs. By capturing the resulting feature drift of internal representations and output confidence, our method achieves significantly higher accuracy than prior approaches. 
Experiments demonstrate a strong generalization performance for unseen samples from the same distribution scenarios, where earlier attacks struggle. Our results further suggest that feature-projection drift can serve as a lightweight, model-internal signal for MIA.
This work has practical implications for auditing and privacy: it enables reliable verification of whether specific data was used in training, aiding data creators and auditors in ensuring transparency. At the same time, it highlights a significant privacy risk, emphasizing the need for effective privacy protections during model development.

%Bibliography
\bibliographystyle{unsrt}  
\bibliography{references}  

@article{Shokri2017,
    author = {Shokri, Reza and Stronati, Marco and Song, Congzheng and Shmatikov, Vitaly},
    title = {Membership Inference Attacks Against Machine Learning Models},
    journal = {IEEE Symposium on Security and Privacy},
    year = {2017},
    pages = {3--18}
}

@inproceedings{Yeom2018,
    author = {Yeom, Samuel and Giacomelli, Irene and Fredrikson, Matt and Jha, Somesh},
    title = {Privacy Risk in Machine Learning: Analyzing the Connection to Overfitting},
    booktitle = {IEEE Computer Security Foundations Symposium},
    year = {2018}
}

@article{Maini2024,
  title={On the difficulty of membership inference attacks},
  author={Rezaei, Shahbaz and Liu, Xin},
  journal={Proceedings of the IEEE/CVF Conference on Computer Vision and Pattern Recognition},
  pages={7892--7900},
  year={2021}
}

@inproceedings{Carlini2022,
  title={Extracting training data from large language models},
  author={Carlini, Nicholas and Tramer, Florian and Wallace, Eric and Jagielski, Matthew and Herbert-Voss, Ariel and Lee, Katherine and Roberts, Adam and Brown, Tom and Song, Dawn and Erlingsson, Ulfar and others},
  booktitle={30th USENIX security symposium (USENIX Security 21)},
  pages={2633--2650},
  year={2021}
}

@inproceedings{Nasr2019,
    author = {Nasr, Milad and Shokri, Reza and Houmansadr, Amir},
    title = {Comprehensive Privacy Analysis of Deep Learning: Passive and Active White-box Inference Attacks against Centralized and Federated Learning},
    booktitle = {IEEE Symposium on Security and Privacy},
    year = {2019}
}

@article{Elhage2022,
    author = {Elhage, Nelson and Nanda, Neel and Olsson, Catherine and Henighan, Tom and Joseph, Nicholas and Mann, Ben and Askell, Amanda and Bai, Yuntao and Chen, Anna and Conerly, Tom and others},
    title = {Toy Models of Superposition},
    journal = {arXiv preprint arXiv:2209.10652},
    year = {2022}
}

@article{Bricken2023,
  title={Towards monosemanticity: Decomposing language models with dictionary learning},
  author={Bricken, Trenton and Templeton, Adly and Batson, Joshua and Chen, Brian and Jermyn, Adam and Conerly, Tom and Turner, Nick and Anil, Cem and Denison, Carson and Askell, Amanda and others},
  journal={Transformer Circuits Thread},
  volume={2},
  year={2023}
}

@article{Gong2025,
    author = {Gong, Neil Zhenqiang and Zhang, Yuheng and Chen, Xinyun},
    title = {Exploiting Polysemantic Neurons for Adversarial Interventions in Language Models},
    journal = {IEEE Symposium on Security and Privacy},
    year = {2025}
}

@article{Choquette2021,
    author = {Choquette-Choo, Christopher A. and Tramer, Florian and Carlini, Nicholas and Papernot, Nicolas},
    title = {Label-Only Membership Inference Attacks},
    journal = {International Conference on Machine Learning},
    year = {2021}
}

@article{Touvron2023,
  title={Llama: Open and efficient foundation language models},
  author={Touvron, Hugo and Lavril, Thibaut and Izacard, Gautier and Martinet, Xavier and Lachaux, Marie-Anne and Lacroix, Timoth{\'e}e and Rozi{\`e}re, Baptiste and Goyal, Naman and Hambro, Eric and Azhar, Faisal and others},
  journal={arXiv preprint arXiv:2302.13971},
  year={2023}
}

@article{GPTNeo2021,
  author  = "Black, Sidney and others",
  year    = 2021,
  title   = "{GPT-Neo: Large Scale Autoregressive Language Modeling with Mesh-Tensorflow}",
  journal = "EleutherAI Blog",
  note    = "\\url{https://www.eleuther.ai/projects/gpt-neo/}"
}

@inproceedings{shi2023min_k,
  title={Detecting pretraining data from large language models},
  author={Shi, Weijia and Ajith, Anirudh and Xia, Mengzhou and Huang, Yangsibo and Liu, Daogao and Blevins, Terra and Chen, Danqi and Zettlemoyer, Luke},
   booktitle={arXiv preprint arXiv:2310.16789},
  year={2023}
}

@inproceedings{fu2022spvmia,
  author    = "Fu, Qian and Li, Haoyang and Xu, Xinyun and Li, Peizhi and Gong, Neil Zhenqiang and Zhang, Xinyu and Song, Dawn Xiaodong",
  title     = "{Label-Only Membership Inference Attacks Against Large Language Models}",
  booktitle = "Proceedings of the 38th Annual Computer Security Applications Conference (ACSAC '22)",
  year      = 2022,
  pages     = "1096--1110",
  address   = "Austin, TX",
  publisher = "ACM"
}

@inproceedings{mattern2023neighbormia,
  author    = "Mattern, Daniel and Geiping, Jonas and Goldblum, Micah and Goldstein, Tom",
  title     = "{Membership Inference Attacks and Defenses in the Wild}",
  booktitle = "International Conference on Learning Representations (ICLR)",
  year      = 2023
}

@misc{shi2023detecting,
  title={Detecting Pretraining Data from Large Language Models},
  author={Weijia Shi and Anirudh Ajith and Mengzhou Xia and Yangsibo Huang and Daogao Liu and Terra Blevins and Danqi Chen and Luke Zettlemoyer},
  year={2023},
  eprint={2310.16789},
  archivePrefix={arXiv},
  primaryClass={cs.CL},
  note={\url{https://huggingface.co/datasets/swj0419/WikiMIA}}
}

@inproceedings{karamolegkou2023,
  title = {Copyright {{Violations}} and {{Large Language Models}}},
  booktitle = {Proceedings of the 2023 {{Conference}} on {{Empirical Methods}} in {{Natural Language Processing}}},
  author = {Karamolegkou, Antonia and Li, Jiaang and Zhou, Li and S{\o}gaard, Anders},
  year = {2023},
  pages = {7403--7412},
  urldate = {2025-07-31}
}

@misc{freeman2024,
  title = {Exploring {{Memorization}} and {{Copyright Violation}} in {{Frontier LLMs}}: {{A Study}} of the {{New York Times}} v. {{OpenAI}} 2023 {{Lawsuit}}},
  shorttitle = {Exploring {{Memorization}} and {{Copyright Violation}} in {{Frontier LLMs}}},
  author = {Freeman, Joshua and Rippe, Chloe and Debenedetti, Edoardo and Andriushchenko, Maksym},
  year = {2024},
  month = dec,
  number = {arXiv:2412.06370},
  eprint = {2412.06370},
  primaryclass = {cs},
  publisher = {arXiv},
  doi = {10.48550/arXiv.2412.06370},
  urldate = {2025-07-31},
}

@article{mueller2024,
  title = {{{LLMs}} and {{Memorization}}: {{On Quality}} and {{Specificity}} of {{Copyright Compliance}}},
  shorttitle = {{{LLMs}} and {{Memorization}}},
  author = {Mueller, Felix B. and G{\"o}rge, Rebekka and Bernzen, Anna K. and Pirk, Janna C. and Poretschkin, Maximilian},
  year = {2024},
  journal = {Proceedings of the AAAI/ACM Conference on AI, Ethics, and Society},
  volume = {7},
  number = {1},
  pages = {984--996},
  issn = {3065-8365},
  doi = {10.1609/aies.v7i1.31697},
  copyright = {Copyright (c) 2024 Association for the Advancement of Artificial Intelligence},
}

@article{huhongsheng2022,
  title = {Membership {{Inference Attacks}} on {{Machine Learning}}: {{A Survey}}},
  shorttitle = {Membership {{Inference Attacks}} on {{Machine Learning}}},
  author = {Hu, Hongsheng and Salcic, Zoran and Sun, Lichao and Dobbie, Gillian and Yu, Philip S. and Zhang, Xuyun},
  year = {2022},
  month = sep,
  journal = {ACM Computing Surveys (CSUR)},
  publisher = {ACMPUB27New York, NY},
  doi = {10.1145/3523273},
}

@article{niu2024,
  title = {A Survey on Membership Inference Attacks and Defenses in Machine Learning},
  author = {Niu, Jun and Liu, Peng and Zhu, Xiaoyan and Shen, Kuo and Wang, Yuecong and Chi, Haotian and Shen, Yulong and Jiang, Xiaohong and Ma, Jianfeng and Zhang, Yuqing},
  year = {2024},
  month = sep,
  journal = {Journal of Information and Intelligence},
  volume = {2},
  number = {5},
  pages = {404--454},
  issn = {2949-7159},
  doi = {10.1016/j.jiixd.2024.02.001},
}

@inproceedings{meeus2025,
  title = {{{SoK}}: {{Membership Inference Attacks}} on {{LLMs}} Are {{Rushing Nowhere}} (and {{How}} to {{Fix It}})},
  shorttitle = {{{SoK}}},
  booktitle = {2025 {{IEEE Conference}} on {{Secure}} and {{Trustworthy Machine Learning}} ({{SaTML}})},
  author = {Meeus, Matthieu and Shilov, Igor and Jain, Shubham and Faysse, Manuel and Rei, Marek and {de Montjoye}, Yves-Alexandre},
  year = {2025},
  month = apr,
  pages = {385--401},
  doi = {10.1109/SaTML64287.2025.00028},
  urldate = {2025-08-01}
}

@article{yan2024protecting,
  title={On protecting the data privacy of large language models (llms): A survey},
  author={Yan, Biwei and Li, Kun and Xu, Minghui and Dong, Yueyan and Zhang, Yue and Ren, Zhaochun and Cheng, Xiuzhen},
  journal={arXiv preprint arXiv:2403.05156},
  year={2024}
}

@article{hu2022membership,
  title     = {Membership Inference Attacks on Machine Learning: A Survey},
  author    = {Hu, Hongsheng and Tang, Jie and Cai, Zhipeng},
  journal   = {IEEE Transactions on Knowledge and Data Engineering},
  volume    = {34},
  number    = {8},
  pages     = {4010--4029},
  year      = {2022},
  doi       = {10.1109/TKDE.2021.3087682},
  publisher = {IEEE}
}

@article{aubinais2025membership,
  title={Membership Inference Risks in Quantized Models: A Theoretical and Empirical Study},
  author={Aubinais, Eric and Formont, Philippe and Piantanida, Pablo and Gassiat, Elisabeth},
  journal={arXiv preprint arXiv:2502.06567},
  year={2025}
}

@inproceedings{shi2023membership,
  title     = {Membership Inference Attack against Language Models via Model Adaptation},
  author    = {Shi, Xuan and Song, Linqi and Qu, Lijun},
  booktitle = {Findings of the Association for Computational Linguistics: ACL 2023},
  pages     = {6105--6116},
  year      = {2023},
  publisher = {Association for Computational Linguistics}
}

@article{mattern2023membership,
  title     = {Membership Inference Attacks via Neighbourhood Analysis},
  author    = {Mattern, Christoph and Kaundinya, Sai Suman and Kairouz, Peter and Song, Dawn},
  journal   = {arXiv preprint arXiv:2305.15885},
  year      = {2023}
}

@inproceedings{Leino2020,
  title     = {Stolen Memories: Leveraging Model Memorization for Calibrated White-Box Membership Inference},
  author    = {Leino, Karl and Fredrikson, Matt},
  booktitle = {29th USENIX Security Symposium},
  pages     = {1605--1622},
  year      = {2020},
  publisher = {USENIX Association}
}

@article{duan2024membership,
  title={Do membership inference attacks work on large language models?},
  author={Duan, Michael and Suri, Anshuman and Mireshghallah, Niloofar and Min, Sewon and Shi, Weijia and Zettlemoyer, Luke and Tsvetkov, Yulia and Choi, Yejin and Evans, David and Hajishirzi, Hannaneh},
  journal={arXiv preprint arXiv:2402.07841},
  year={2024}
}

@article{fu2024selfprompt,
  title={Membership inference attacks against fine-tuned large language models via self-prompt calibration},
  author={Fu, Wenjie and Wang, Huandong and Gao, Chen and Liu, Guanghua and Li, Yong and Jiang, Tao},
  journal={Advances in Neural Information Processing Systems},
  volume={37},
  pages={134981--135010},
  year={2024}
}

@article{abadi2016deep,
  title     = {Deep Learning with Differential Privacy},
  author    = {Abadi, Mart{\'\i}n and Chu, Andy and Goodfellow, Ian and McMahan, H. Brendan and Mironov, Ilya and Talwar, Kunal and Zhang, Li},
  journal   = {Proceedings of the 2016 ACM SIGSAC Conference on Computer and Communications Security (CCS)},
  pages     = {308--318},
  year      = {2016},
  publisher = {ACM}
}

@article{tonni2020data,
  title={Data and model dependencies of membership inference attack},
  author={Tonni, Shakila Mahjabin and Vatsalan, Dinusha and Farokhi, Farhad and Kaafar, Dali and Lu, Zhigang and Tangari, Gioacchino},
  journal={arXiv preprint arXiv:2002.06856},
  year={2020}
}

@inproceedings{hinder2022suitability,
  title={Suitability of different metric choices for concept drift detection},
  author={Hinder, Fabian and Vaquet, Valerie and Hammer, Barbara},
  booktitle={International Symposium on Intelligent Data Analysis},
  pages={157--170},
  year={2022},
  organization={Springer}
}

@article{xu2024tracking,
  title={Tracking the feature dynamics in llm training: A mechanistic study},
  author={Xu, Yang and Wang, Yi and Huang, Hengguan and Wang, Hao},
  journal={arXiv preprint arXiv:2412.17626},
  year={2024}
}

@inproceedings{tirumala2022memorization,
  title     = {Memorization Without Overfitting: Analyzing the Training Dynamics of Large Language Models},
  author    = {Tirumala, Kushal and Markosyan, Aram H. and Zettlemoyer, Luke and Aghajanyan, Armen},
  booktitle = {Advances in Neural Information Processing Systems},
  year      = {2022}
}

@inproceedings{he2025towards,
  title={Towards label-only membership inference attack against pre-trained large language models},
  author={He, Yu and Li, Boheng and Liu, Liu and Ba, Zhongjie and Dong, Wei and Li, Yiming and Qin, Zhan and Ren, Kui and Chen, Chun},
  booktitle={USENIX Security},
  year={2025}
}

@inproceedings{hayesexploring,
  title={Exploring the limits of strong membership inference attacks on large language models},
  author={Hayes, Jamie and Shumailov, Ilia and Choquette-Choo, Christopher A and Jagielski, Matthew and Kaissis, Georgios and Nasr, Milad and Annamalai, Meenatchi Sundaram Muthu Selva and Mireshghallah, Niloofar and Shilov, Igor and Meeus, Matthieu and others},
  booktitle={The 39th Annual Conference on Neural Information Processing Systems},
   year={2025}
}

@inproceedings{carlini2022membership,
  title={Membership inference attacks from first principles},
  author={Carlini, Nicholas and Chien, Steve and Nasr, Milad and Song, Shuang and Terzis, Andreas and Tramer, Florian},
  booktitle={2022 IEEE symposium on security and privacy (SP)},
  pages={1897--1914},
  year={2022},
  organization={IEEE}
}

@article{team2024gemma,
  title={Gemma: Open models based on {Gemini} research and technology},
  author={Team, Gemma and Mesnard, Thomas and Hardin, Cassidy and Dadashi, Robert and Bhupatiraju, Surya and Pathak, Shreya and Sifre, Laurent and Rivi{\`e}re, Morgane and Kale, Mihir Sanjay and Love, Juliette and others},
  journal={arXiv preprint arXiv:2403.08295},
  year={2024}
}

@article{maini2024tofu,
  title={Tofu: A task of fictitious unlearning for {LLMs}},
  author={Maini, Pratyush and Feng, Zhili and Schwarzschild, Avi and Lipton, Zachary C and Kolter, J Zico},
  journal={arXiv preprint arXiv:2401.06121},
  year={2024}
}

@inproceedings{ibanez2025lumia,
  title={{LUMIA}: Linear probing for Unimodal and MultiModal Membership Inference Attacks leveraging internal LLM states},
  author={Ibanez-Lissen, Luis and Gonzalez-Manzano, Lorena and de Fuentes, Jose Maria and Anciaux, Nicolas and Garcia-Alfaro, Joaquin},
  booktitle={European Symposium on Research in Computer Security},
  pages={186--206},
  year={2025},
  organization={Springer}
}

@article{ranjan2026razor,
  title={RAZOR: Ratio-Aware Layer Editing for Targeted Unlearning in Vision Transformers and Diffusion Models},
  author={Ranjan, Ravi and Grover, Utkarsh and Lin, Xiaomin and Polyzou, Agoritsa},
  journal={arXiv preprint arXiv:2603.14819},
  year={2026}
}

@article{ranjan2026catrag,
  title={CatRAG: Functor-Guided Structural Debiasing with Retrieval Augmentation for Fair LLMs},
  author={Ranjan, Ravi and Grover, Utkarsh and Akewar, Mayur and Lin, Xiaomin and Polyzou, Agoritsa},
  journal={arXiv preprint arXiv:2603.21524},
  year={2026}
}

@article{ranjan2026position,
  title={Position: Llms must use functor-based and rag-driven bias mitigation for fairness},
  author={Ranjan, Ravi and Grover, Utkarsh and Polyzou, Agorista},
  journal={arXiv preprint arXiv:2603.07368},
  year={2026}
}

@article{kumar2024trustworthiness,
  title={Trustworthiness of llms in medical domain},
  author={Kumar, Ravi R and Pramanik, Vishal and Grover, Utkarsh and Ganapam, Venkata Ramesh},
  journal={Researchgate preprint},
  year={2024}
}

@article{grover2026embodied,
  title={Embodied Foundation Models at the Edge: A Survey of Deployment Constraints and Mitigation Strategies},
  author={Grover, Utkarsh and Ranjan, Ravi and Mao, Mingyang and Dong, Trung Tien and Praveen, Satvik and Wu, Zhenqi and Chang, J Morris and Mohsenin, Tinoosh and Sheng, Yi and Polyzou, Agoritsa and others},
  journal={arXiv preprint arXiv:2603.16952},
  year={2026}
}

\appendix
\clearpage
\section*{Appendix}

\section{Pseudo code}
\label{app:pseudo}

\begin{algorithm}[h!]
\centering
\caption{G-Drift MIA}
\label{alg:gdrift}
\begin{algorithmic}[1]
  \REQUIRE JSON link \texttt{json\_link} to dataset of $(q,a,\ell)$
  \ENSURE Trained logistic‐regression MIA classifier and its accuracy/ROC‐AUC
  \STATE \textbf{Constants:} $\mathrm{MODEL\_NAME}\gets\texttt{"llm-model"}$, $\eta\gets10^{-2}$
  \STATE $D\gets\textsc{LoadDataset}(\texttt{json\_link})$
  \STATE Initialize tokenizer: 
    \begin{ALC@g}
      \STATE $\quad \text{tokenizer}\gets\text{AutoTokenizer.from\_pretrained}(\mathrm{MODEL\_NAME})$
    \end{ALC@g}
  \STATE Initialize model:
    \begin{ALC@g}
      \STATE $\quad \text{model}\gets\text{AutoModelForCausalLM.from\_pretrained}(\mathrm{MODEL\_NAME},\,\text{output\_hidden\_states}=\text{True})$
    \end{ALC@g}
  \STATE $\text{model.train}()$
  \STATE $\Theta_{\rm orig}\gets\textsc{CloneParams}(\text{model})$
  \STATE Sample random unit vector $v\in\mathbb{R}^d$
  \STATE $\mathcal{F}\gets\emptyset$
  \FORALL{$(q,a,\ell)\in D$}
    \STATE $x\gets\text{tokenizer}(q)$
    \STATE $\tau\gets\textsc{FirstSubtokenID}(a)$
    \STATE $(\text{logits}_0, h_0)\gets\text{model.forward}(x)$
    \STATE $\text{loss}_0\gets\mathrm{CE}(\text{logits}_0,\tau)$
    \STATE $\text{logit}_0\gets \text{logits}_0[\tau]$
    \STATE $\text{feat}_0\gets h_0\cdot v$
    \STATE $\text{optimizer}\gets\text{SGD}(\text{model.params}(),\text{lr}=\eta)$
    \STATE $\text{optimizer.zero\_grad}()$
    \STATE $\text{loss}_u\gets\mathrm{CE}(\text{model}(x).\text{logits},\tau)$
    \STATE $(-\text{loss}_u).\text{backward}()$
    \STATE $\text{optimizer.step}()$
    \STATE $(\text{logits}_1,h_1)\gets\text{model.forward}(x)$
    \STATE $\text{loss}_1\gets\mathrm{CE}(\text{logits}_1,\tau)$
    \STATE $\text{logit}_1\gets \text{loogits}_1[\tau]$
    \STATE $\text{feat}_1\gets h_1\cdot v$
    \STATE $\text{drift}\gets \|h_0 - h_1\|$
    \STATE Append $([\text{loss}_0,\text{logit}_0,\text{feat}_0,\text{loss}_1,\text{logit}_1,\text{feat}_1,\text{drift}],\,\ell)$ to $\mathcal{F}$
    \STATE $\textsc{RestoreParams}(\text{model},\Theta_{\rm orig})$
  \ENDFOR
  \STATE Split $\mathcal{F}$ into train/test sets $(X_{\rm tr},y_{\rm tr},X_{\rm te},y_{\rm te})$
  \STATE $\mathrm{clf}\gets\text{LogisticRegression}()$
  \STATE $\mathrm{clf.fit}(X_{\rm tr},y_{\rm tr})$
  \STATE $\hat y\gets\mathrm{clf.predict}(X_{\rm te})$
  \STATE $\hat p\gets\mathrm{clf.predict\_proba}(X_{\rm te})[:,1]$
  \STATE $\mathrm{accuracy}\gets\mathrm{accuracy\_score}(y_{\rm te},\hat y)$
  \STATE $\mathrm{AUC}\gets\mathrm{roc\_auc\_score}(y_{\rm te},\hat p)$
  \STATE \textbf{return} $\mathrm{accuracy},\,\mathrm{AUC}$
\end{algorithmic}
\end{algorithm}

\section{Detailed Experimental Settings}
\label{app:exp-setting}
In evaluating the Gradient-Induced Feature Drift (G-Drift) method, we conducted experiments using three modern transformer-based language models \textbf{Llama-3}, \textbf{Gemma-3}, and \textbf{GPT-Neo-2} across the three subsets of the \textbf{WikiMIA benchmark}: \emph{World Facts}, \emph{Real Authors}, and \emph{Books}. Models were fine-tuned on the corresponding training subsets, while non-member data consisted of Replica Q\&A pairs (structurally similar but content-distinct from training data) and narrative excerpts from the Books portion of the benchmark.

\textbf{G-Drift Method.}
Our method applies a single gradient-ascent step with a learning rate of $10^{-2}$ to intentionally perturb model parameters in the direction of increasing loss on a queried instance. For each sample, we record both pre- and post-update statistics: logits, loss values, hidden state activations, and the induced feature drift measured via the Euclidean norm $\lVert \Delta h \rVert$. A logistic regression classifier is trained on these drift features to distinguish members from non-members.

We compare G-Drift against a suite of strong membership-inference baselines spanning black-box, reference-based, shadow-model, and label-only threat models:

\begin{itemize}

\item \textbf{Probability Threshold (Min-k\%).}
This black-box attack estimates the percentile rank of the true likelihood $p(y \mid x)$ within a reference distribution~\cite{shi2023min_k}. Membership is predicted when $p(y \mid x)$ exceeds a $k^\text{th}$-percentile threshold. We compute token-level negative log-likelihood and calibrate thresholds using the median training-set score.

\item \textbf{Shadow Model Attack (Neighbour-MIA).}
Following Mattern et al.~\cite{mattern2023neighbormia}, Neighbour-MIA trains shadow models on auxiliary data that approximate the target model's behavior. Membership is inferred by comparing the queried instance's outputs to outputs observed during shadow-model training. In our implementation, neighbourhood variants are generated using a masked language model (``bert-base-uncased’’), and membership is inferred from the difference between original losses and averaged neighbour losses.

\item \textbf{Perplexity-Based Likelihood (Perplexity-PL) and Zlib.}
Based on Carlini et al.~\cite{Carlini2022}, Perplexity-PL uses sequence-level perplexity as the membership statistic. The Zlib variant incorporates compression entropy by dividing perplexity by the compressed bit length. Thresholds are optimized using training data to detect samples that are unusually easy for the model to predict or compress.

\item \textbf{Similarity-Based Attack (SPV-MIA).}
SPV-MIA~\cite{fu2022spvmia} assumes access to reference data drawn from the same distribution and measures embedding-space similarity between the query’s outputs and those of reference samples. Membership is inferred when similarity exceeds a calibrated threshold. Our implementation monitors representation stability until convergence before thresholding cosine similarities.

\item \textbf{PETAL (Label-Only MIA).}
PETAL~\cite{he2025towards} operates under a stricter label-only setting in which logits are unavailable. It infers membership by computing semantic similarity between model-predicted tokens and reference answers, serving as a strong baseline in restricted-access scenarios.

\end{itemize}

All methods use an 70/10/20 train–validation-test split and logistic regression classifiers with L2 regularization selected via cross-validation. AUC is used as the primary evaluation metric to ensure consistent and rigorous comparison across all approaches.

\subsection{Supporting Methodological Context}
\label{app:supporting_context}

G\textsc{-}Drift is situated within a broader research program on trustworthy, controllable, and deployable foundation models. In particular, our use of a \emph{targeted gradient-based intervention} is conceptually aligned with recent work on selective model editing and unlearning, where carefully localized updates are used to remove or suppress specific behaviors while preserving general utility \cite{ranjan2026razor}. From a safety and fairness perspective, this view also resonates with recent efforts to modify model behavior through structure-aware debiasing and retrieval-grounded correction, which emphasize that reliable intervention requires both principled internal transformations and external grounding \cite{ranjan2026catrag,ranjan2026position}. More broadly, the motivation for auditing internal model responses is consistent with prior work on trustworthiness in high-stakes domains such as medicine, where interpretability and robust diagnostic signals are essential for responsible deployment \cite{kumar2024trustworthiness}. Finally, as foundation models are increasingly deployed in constrained and real-world environments, system-level reliability and controllability become inseparable from model auditing, further motivating lightweight yet informative probes such as gradient-induced feature drift \cite{grover2026embodied}.

These works are not direct baselines for G\textsc{-}Drift, but they provide useful methodological support for our central premise: small, structured interventions can reveal meaningful properties of internal representations, and such signals are valuable for auditing safety, fairness, privacy, and deployment readiness. In this sense, G\textsc{-}Drift contributes a complementary perspective focused specifically on \emph{membership-sensitive representation dynamics} in LLMs.

\section{Quantitative Analysis}
\label{app:quant}

\begin{table}[t!]
\centering
\small
\setlength{\tabcolsep}{5pt}
\renewcommand{\arraystretch}{1.05}
\caption{Drift consistency across semantically similar prompts (illustrative example on LLaMA-3). Members show stable, repeatable feature-projection drift across paraphrases, while non-members exhibit smaller and less consistent drift.}
\label{tab:drift_consistency_paraphrase}
\begin{tabular}{p{3.8cm} c r r r}
\toprule
\textbf{Prompt (paraphrase)} & \textbf{Class} & $\alpha_{\text{before}}$ & $\alpha_{\text{after}}$ & $|\Delta \alpha|$ \\
\midrule
Where would you find the Eiffel Tower? & Member & 2.914 & 3.900 & 0.986 \\
The Eiffel Tower is located in which city? & Member & 2.890 & 3.875 & 0.985 \\
Which city is home to the Eiffel Tower? & Member & 2.905 & 3.890 & 0.985 \\
\midrule
Where would you find the Eiffel Tower? & Non-member & -0.103 & -0.057 & 0.046 \\
The Eiffel Tower is located in which city? & Non-member & -0.120 & -0.030 & 0.090 \\
Which city is home to the Eiffel Tower? & Non-member & -0.085 & -0.010 & 0.075 \\
\bottomrule
\end{tabular}
\end{table}

In Table~\ref{tab:drift_consistency_paraphrase}, we evaluate a setting where the same question–answer pair is included in the training data of the member-class LLMs but excluded from the non-member models, and then apply paraphrased variants of the question to examine how feature drift differs between member and non-member behavior. Table~\ref{tab:drift_consistency_paraphrase} shows that feature projection drift remains highly consistent across semantically equivalent prompts for member samples, while non-members exhibit smaller and more variable drift. This stability indicates that memorized facts are anchored in robust internal feature representations, whereas unseen samples lack such geometric consistency. The result highlights that G\textsc{-}Drift captures semantic memorization rather than prompt-specific artifacts.

\end{document}